\documentclass[10pt,twocolumn,letterpaper]{article}
\usepackage{cvpr}              

\usepackage{graphicx}
\usepackage{amsmath}
\usepackage{amssymb}
\usepackage{booktabs}
\usepackage{microtype}      
\usepackage{xcolor}         
\usepackage{graphicx}
\usepackage{multirow}
\usepackage{bm}
\usepackage{caption}

\definecolor{cvprblue}{rgb}{0.21,0.49,0.74}
\usepackage[pagebackref,breaklinks,colorlinks,citecolor=cvprblue]{hyperref}

\def\paperID{5735} 
\def\confName{CVPR}
\def\confYear{2024}

\title{Evidential Active Recognition: Intelligent and Prudent\\Open-World Embodied Perception}

\author{
Lei Fan\textsuperscript{1},
Mingfu Liang\textsuperscript{1}, 
Yunxuan Li\textsuperscript{1},
Gang Hua\textsuperscript{2} and Ying Wu\textsuperscript{1}\\
\textsuperscript{1}Northwestern University \textsuperscript{2}Wormpex AI Research\\
{\tt\footnotesize \{leifan,mingfuliang2020,yunxuanli2019\}@u.northwestern.edu,ganghua@gmail.com,yingwu@northwestern.edu}\\
}

\begin{document}
\maketitle

\begin{abstract}
Active recognition enables robots to intelligently explore novel observations, thereby acquiring more information while circumventing undesired viewing conditions.
Recent approaches favor learning policies from simulated or collected data, wherein appropriate actions are more frequently selected when the recognition is accurate.
However, most recognition modules are developed under the closed-world assumption, which makes them ill-equipped to handle unexpected inputs, such as the absence of the target object in the current observation.
To address this issue, we propose treating active recognition as a sequential evidence-gathering process, providing by-step uncertainty quantification and reliable prediction under the evidence combination theory.
Additionally, the reward function developed in this paper effectively characterizes the merit of actions when operating in open-world environments.
To evaluate the performance, we collect a dataset from an indoor simulator, encompassing various recognition challenges such as distance, occlusion levels, and visibility.
Through a series of experiments on recognition and robustness analysis, we demonstrate the necessity of introducing uncertainties to active recognition and the superior performance of the proposed method.
\end{abstract}

\vspace{-4pt}
\section{Introduction}
\label{sec:intro}
Passive visual recognition, encompassing a broad range of approaches such as image-based and video-based techniques~\cite{yue2015beyond,he2016deep,dosovitskiy2020image,liu2022video}, has experienced tremendous success in recent decades.
Nonetheless, a contrasting approach, the active recognition system~\cite{aloimonos1988active,aloimonos1990purposive,ammirato2017dataset}, offers unique advantages.
Namely, it benefits from the ability to move within and perceive its environment, allowing it to dynamically determine visual inputs, as opposed to relying on pre-captured images.


\begin{figure}[t]
    \centering
    \includegraphics[width=1\linewidth]{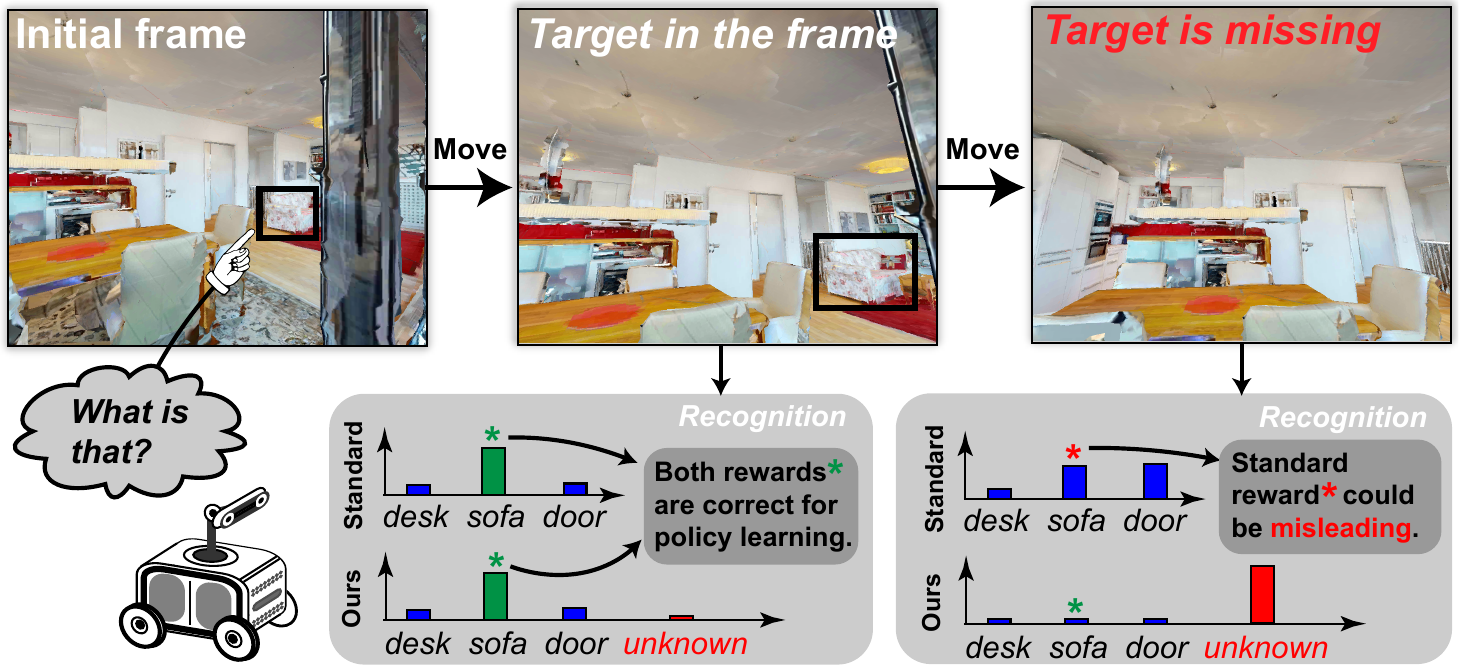}
    \caption{An overview of active recognition agents operating in an open-world environment with different recognition models. The agent is initially queried to perform recognition. In standard practice, the recognition model is unable to identify its own failures, especially when confronted with unexpected inputs, which may lead to incorrect rewards for the policy. In contrast, our approach estimates uncertainty at each step and fuses the collected evidence to provide a final prediction.}
    \vspace{-9pt}
    \label{fig:overview}
\end{figure}
A number of approaches~\cite{andreopoulos2013computational,atanasov2014nonmyopic,kasaei2018perceiving,bajcsy2018revisiting} in active recognition have been proposed over the years with information gain quantification~\cite{denzler2002information}, auxiliary task modeling~\cite{cheng2018geometry,ramakrishnan2018sidekick}, and discriminative feature discovery~\cite{jayaraman2016look}.
Given its sequential nature, prevalent approaches~\cite{jayaraman2016look,ding2023learning,yang2019embodied,fan2021flar} involve combining recognition models, such as object detectors or classifiers, with reinforcement learning techniques.
The rewards, which can take the form of recognition accuracy, serve as incentives for the policy and are directly associated with the outputs of the recognition model.
Nevertheless, the reward could be misleading when deploying the agent in open environments~\cite{gallos2019active}.
Consider an agent being queried with a target in the initial observation. 
The target may be out of view during subsequent movements, \ie, an unexpected input for the recognition system.
As a consequence, the recognition might fail to provide rewards that accurately represent the worth of actions being taken, impeding effective policy learning.

Regarding this challenge, our insight is derived from an instinctive understanding of how humans learn from experience. 
We consciously weigh our learning based on the level of trust we have in our observations. 
When faced with a high degree of uncertainty, we exercise caution in drawing conclusions from our experiences.
Likewise, the agent should evaluate the uncertainty in its observations to avoid being negatively influenced by unexpected inputs, such as when the target object exhibits high ambiguity or is simply not present in the input.

Besides, compared to passive recognition, the importance of modeling uncertainty is more prominent in active recognition, since the operating environment for agents is inherently open and unpredictable.
This area, however, remains under-explored as recent methods often resort to recognition models built under the closed-world assumption and then directly using softmax outputs as the confidence measure.
In this paper, we choose to model active recognition as a sequential evidence-gathering process, wherein the agent learns to explore trustworthy observations and gives the final prediction based on a collection of opinions.
The comparison of the proposed method with standard active recognition methods is depicted in Figure.~\ref{fig:overview}, where our agent can assign high uncertainties to inputs lacking target objects.
After actively exploring multiple views, the final prediction of the target is derived from a combination of collected opinions using Dempster-Shafer's combination theory~\cite{josang2016subjective}.

Furthermore, to evaluate active recognition approaches, we collect a new dataset from an existing indoor simulator~\cite{szot2021habitat,savva2019habitat} where the target could be queried using a bounding box.
In addition, each test instance in our dataset is accompanied by a recognition difficulty level, taking into account factors such as visibility, observed pixels, and distance.
We expect that the difficulty level will more effectively demonstrate the value of active recognition during evaluation, as the agent can potentially overcome these challenges, unlike in passive recognition.

The primary contributions of this paper are as follows: (1) We develop a novel active recognition approach using evidence theory to address the challenges of open-world environments, which are largely overlooked by other methods. (2) We build a new dataset for evaluating active recognition agents, encompassing 13200 test instances across 27 different indoor object categories. Importantly, each test instance is assigned a recognition difficulty level for a more comprehensive assessment of methods. (3) We conduct extensive experiments and analysis of the proposed method and other baselines. These investigations demonstrate the trustworthiness and superior performance of our method.


\section{Related Work}
\label{sec:related_work}
\noindent{\bf{Active recognition.}}
Embodied artificial intelligence, a long-standing field driven by enabling agents to learn through interactions, has been investigated across various streams~\cite{chaplot2020object,gadre2022continuous,kotar2022interactron,nilsson2021embodied}. 
As a notable branch, active recognition~\cite{aloimonos1990purposive,ballard1991animate,ammirato2017dataset,fang2020move,cheng2018geometry,gallos2019active,liu2018extreme} lets the agent explore observations with its own incentives, thereby achieving improved recognition performance.

Recent approaches~\cite{jayaraman2016look,fan2021flar,ding2023learning,yang2019embodied} in active recognition typically represent the recognition as a Markov Decision Process and rely on Reinforcement Learning (\textit{RL}) to learn moving strategies from interactions.
Unlike other \textit{RL} scenarios where the reward function is explicitly specified by the environment~\cite{mnih2016asynchronous,schulman2017proximal}, the reward function in active recognition is associated with the recognition module, which is defined as information gain~\cite{liu2018extreme}, the recognition accuracy~\cite{jayaraman2016look,jayaraman2018end,cheng2018geometry,yang2019embodied}, or the recognition score~\cite{ammirato2017dataset,ding2023learning}. 
In~\cite{jayaraman2016look,ramakrishnan2019emergence}, the authors propose an active recognition agent with modules to aggregate historical information and predict the next views. The policy learning is supervised by a binary reward function checking whether the top prediction is correct.
In parallel, \cite{ding2023learning} selects the surrogate reward, defined as the detection score, during policy learning.

However, these reward definitions might not adequately reflect the value of actions when the agent encounters unexpected visual inputs. 
This is because the recognition model was developed within a closed-world setting.
Improper rewards can compromise the recognition policy after training. 





\noindent{\bf{Uncertainty estimation.}} Conventional visual recognition models are unable to identify their own failures. Nonetheless, this ability could be essential when it comes to real-world applications, particularly for embodied agents. Bayesian neural networks~\cite{gal2015bayesian,dusenberry2020efficient,kristiadi2020being,kwon2020uncertainty} equip the recognition models predicting uncertainties as the mutual information between data and parameters. In stark contrast, Evidential Deep Learning~\cite{amini2020deep,sensoy2018evidential} established on the Dempster-Shafer Theory has been investigated in different applications~\cite{bao2021evidential,corbiere2021beyond,han2022trusted,fan2023flexible}, which learn the prior of the categorical prediction directly. Inspired by the success of uncertainty quantification of evidential approaches, the proposed method formulates active recognition as a sequential multi-source evidence fusion under the same frame of discernment. With this modeling, we can more effectively combine knowledge in consecutive observations and handle open-world challenges.


\section{Task Settings and Notations}
\label{sec:task}
We describe our setting by deploying the active object recognition agent in an indoor environment. 

The recognition agent is spawned at a location in an indoor environment without prior knowledge about the scene, {\it{e.g.}}, the map. At the initial timestep $t=1$, the agent is queried with a target $x$ in the current visual observation $v^1$, selected by a bounding box $q^{box}_x$. And a total of $T$ timesteps is allowed for the agent to obtain the final prediction of the target category. Besides the final step, the agent could take an additional action $a^t\in\mathcal{A}$ at $t=1,\dots,T-1$ to change its viewing point, where $\mathcal{A}$ is the action space. By taking movements, the agent could improve its recognition performance by collecting information and give the final predicted label $\hat{y}$ at $t=T$. It is worth noting that the setting could be smoothly generalized to other simulators and different types of targets, like active scene recognition~\cite{xiao2012recognizing,jayaraman2016look,ramakrishnan2021exploration,fan2021flar}.

To provide details about our setup, the observation $v$ is captured by an RGB camera mounted at the height of $1$ meter above the ground, featuring a height-by-width resolution of $640\times800$. 
The action space is defined as $\mathcal{A}=\{{\texttt{move\_forward}}, {\texttt{turn\_left}}, {\texttt{turn\_right}}, {\texttt{look\_up}},\\{\texttt{look\_down}}\}$.
Actions allow the agent to move $0.25$m, turn $10$ degrees, or tilt $10$ degrees.
Previous works on active recognition~\cite{ding2023learning,yang2019embodied} usually do not contain {\texttt{look\_up}} and {\texttt{look\_down}} in to their action space. This may potentially compromise the agent's performance, particularly in the case of objects situated at higher locations in the observation.
As the agent forms multi-round interactions with the environment, the objective of an effective approach lies in three major components, including visual classification, information aggregation, and intelligent moving policies, which we will elaborate on in later sections.


\section{Evidential Active Recognition}
\label{sec:method}
In this section, we introduce a model called Evidential Active Recognition, which is designed to address the challenges associated with developing agents in an open-world context. 
We comprehend active recognition as an evidence-collecting procedure in which the agent explores and gathers knowledge to make predictions.
Evidence for each class is estimated on a per-step basis, complemented by an additional term that describes the uncertainty.
Intuitively, as the agent freely moves in the environment, uncertainty appears when the target is absent, or when the viewpoint is not optimal for acquiring discriminative features.
To capture the uncertainty, the Dirichlet prior is placed for known classes~\cite{sensoy2018evidential} to provide trustworthy results.
In comparison to active recognition approaches that utilize softmax outputs, the estimated uncertainty then plays a key role in policy learning by redefining the reward function.
The final multinomial opinion on the target category follows the Subject Logic~\cite{barnett2008computational,josang2016subjective} to fuse evidence from different views at the last step.

\subsection{Preliminaries}
Current visual recognition models often rely on the softmax operator to give probabilities.
Nonetheless, as the probability is normalized, it could lead to overconfident predictions or even failures when handling unknown inputs. 
Evidential deep learning approaches~\cite{sensoy2018evidential}, on the contrary, choose to develop uncertainties under the scheme towards subjective probabilities~\cite{barnett2008computational,josang2016subjective}. For the next paragraph, we will start by introducing the formulation of evidential deep learning for single-frame prediction.

For a $K$-class recognition task, the frame of discernment $\Theta=\{k, 1\leq k\leq K\}$ contains $K$ exclusive singletons, {\it{e.g.}}, class labels.
Considering the visual observation $v^t$ at timestep $t$, we measure the mass in mutually exclusive propositions with a belief function $b_k^t$ leveraging the Dempster-Shafer Theory of Evidence (DST)~\cite{barnett2008computational}.
By providing an overall uncertainty mass $u^t$, these $K+1$ mass values satisfy
\begin{equation}
    \sum_{k=1}^K b_k^t+u^t=1, \quad 0\leq u^t,b_k^t\leq 1.
    \label{eq:belief_sum}
\end{equation}
The evidence $e^t_k$ is defined as the support evidence collected from the current observation $v^t$ to $k^{th}$ singleton. To form the opinion, the evidence $e^t_k$ is introduced to associate the belief function with Dirichlet distribution parameters $\alpha_k^t$ by
\begin{equation}
    b^t_k=\frac{e^t_k}{S^t} \; {\textnormal{ and }} \; u^t=\frac{K}{S^t},
    \label{eq:belief_unc}
\end{equation}
where $S^t=\sum^K_{k=1}\alpha_k^t=\sum^K_{k=1}(e^t_k+1)$. As the conjugate prior for the multinomial distribution, the Dirichlet distribution characterized by $\alpha^t=[\alpha_1^t,\dots,\alpha_K^t]$ is used to derive the subject opinion, \ie, the belief function as $b_k^t={e_k^t}/{S^t}$. 
Accordingly, the overall uncertainty $u^t$ arises if no significant evidence is being collected from known classes.

From the DST, the masses defined in Equation.~\ref{eq:belief_sum} is a reduction from the general hyper-opinion set $2^\Theta=\{\emptyset, 1, \dots, K, \{1, 2\}, \dots, \Theta\}$ that contains intimidating $2^K$ propositions. As no mass is assigned to the empty-set proposition, \ie, $b_{\emptyset}^t=0$, the uncertainty $u^t$ is consequently the summation of contained non-singleton belief masses. Namely, the overall uncertainty $u^t$ could be interpreted as the sum of conflicting evidence occurring between 2 or more classes, which could be formulated as
\begin{equation}
    u^t=1-\sum_{k=1}^K b_k^t=\sum_{P,P\in2^\Theta,2\leq |P|\leq K} b_P^t,
    \label{eq:uncertainty}
\end{equation}
where we use $P$ as the symbol for any proposition, and it satisfies $0\leq b_P^t\leq 1$.

\subsection{Evidence Fusion}
After discussing belief and uncertainty for a single-step observation, we now proceed to explore how to fuse evidence among multi-frame predictions.
Two primary fusion strategies are employed in active recognition. The first one is early fusion~\cite{yang2019embodied,jayaraman2016look}, which recurrently aggregates temporal visual information at the feature level. The second one is late fusion, which involves techniques such as voting or averaging the softmax outputs~\cite{ding2023learning}.
We propose fusing per-step evidence within the framework of subjective probabilities~\cite{josang2016subjective}, which can be considered as a form of late fusion.

The insights of doing late fusion are three-fold. First, the unexpected visual input can occur at any timestep during a recognition episode, regardless of whether it takes place during policy training or testing. 
Thus, the feature could be contaminated if it is aggregated at an early stage, and difficult to differentiate at which time the recognition uncertainty is introduced.
This, in turn, impedes the provision of reasonable rewards for evaluating the policy.
Second, late fusion better accommodates the assumption in DST that available evidence from different sources should be independently measured.
Third, compared to recurrently fusing temporal features, the late fusion is unaffected by the order of visual observations.
Considering a group of visual observations, the recognition result should remain unchanged by the order in which they are presented.

With no loss of generality, we formulate the evidence combination between any two basic belief functions $b_k^t,b_k^j$ on different observations under the same frame of discernment, \ie, $\Theta$, as:
\begin{equation}
\label{eq:fusion}
\begin{split}
    b_k&=b_k^t\oplus b_k^j=\frac{1}{\sum_{P^t\cap P^j\neq\emptyset}b_{P^t}^t b_{P^j}^j}\sum_{P^t\cap P^j=k} b_{P^t}^t b_{P^j}^j \\
    &=\frac{b^t_k b^j_k + b^t_k u^j + b^j_k u^t}{1-\sum_{i\neq q}b^t_i b^j_q},
\end{split}
\end{equation}
where $P^t$ and $P^j$ are any two propositions in $2^\Theta$.
Similarly, the uncertainty after the combination is defined as $u=u^tu^j/({1-\sum_{i\neq q}b^t_i b^j_q})$, where the denominator serves as a normalization factor, representing the total valid mass.
It should be noted that the combination reduces conflicting evidence as one single term, \ie, approximate with the uncertainty $u^t$ and $u^j$ as in Equation.~\ref{eq:uncertainty}, to lower the computation complexity.
Moreover, this combination guarantees the final prediction to be more strongly influenced by the belief assignment with a lower uncertainty.

Subsequently, given a sequence of $T$ observations collected by the agent, we derive the final belief function by
\begin{equation}
    b_k=b_k^1\oplus b_k^2\oplus \dots \oplus b_k^T.
\end{equation}
The combination inherently adheres to the commutative property, implying that the order of observations does not impact the outcomes. And the final prediction of the category is $\hat{y}=\arg\max_{k}b_k$.

\subsection{Developing Opinions}
In this section, we outline how to develop opinions for the visual recognition model using training data. 
The deep recognition model used in our approach could be generally denoted as a mapping function $f_{\theta}(\cdot)\rightarrow\mathbb{R}$ with a set of parameters, \ie, $\theta$.
The output of the recognition model is obtained directly from the current input as evidence $e_k, 1\leq k\leq K$ after applying a non-negative function, such as an exponential function or sigmoid.
Following evidential deep learning~\cite{sensoy2018evidential}, the training is implemented as the evidence acquisition on a Dirichlet prior, whose loss is formulated as
\begin{equation}
    \mathcal{L}_{edl}^t=\sum^K_{i=1}y_{i}[\log(\sum^K_{j=1}\alpha_{j}^t)-\log(\alpha_{i}^t)],
\end{equation}
where $\mathbf{y}=[y_1, \dots, y_{i}, \dots, y_{K}]^T$ is the one-hot label, and $\bm{\alpha}^t=[\alpha_1^t, \dots,\alpha_{i}^t,\dots, \alpha_{K}^t]$ with $\alpha_{i}^t=e_i^t+1$ are parameters of a Dirichlet distribution $Dir(\cdot|\bm{\alpha})$.
Additionally, a Kullback-Leibler loss is incorporated to promote mutual exclusivity among singleton beliefs, defined as
\begin{equation}
    \mathcal{L}_{kl}^t=\textit{KL}(Dir(\cdot|\tilde{\bm{\alpha}}^t)||Dir(\cdot|\langle 1, \dots, 1\rangle)),
\end{equation}
where $\tilde{\bm{\alpha}}^t=\mathbf{y}+(1-\mathbf{y})\odot\bm{\alpha}^t$, and $\odot$ is for element-wise multiplication. Combining all terms, the total loss for an observation $v^t$ is
\begin{equation}
    \mathcal{L}^t=\mathcal{L}_{edl}^t + \lambda_{kl}\mathcal{L}_{kl}^t,
\end{equation}
where $\lambda_{kl}$ is an annealing weight to gradually increase the effect of $\mathcal{L}_{kl}^t$.

Although the combination rule could provide fused predictions, we opt to train our recognition model using single observations only.
The rationale behind this choice is that if all observations contain the target to be recognized, adding another loss on fused evidence does not alter the optimality of the loss function. Conversely, if the target is missing in any observations, the training may be adversely impacted.

\subsection{Uncertainty-aware Policy Learning}
\begin{figure*}[t]
    \centering
    \includegraphics[width=1\linewidth]{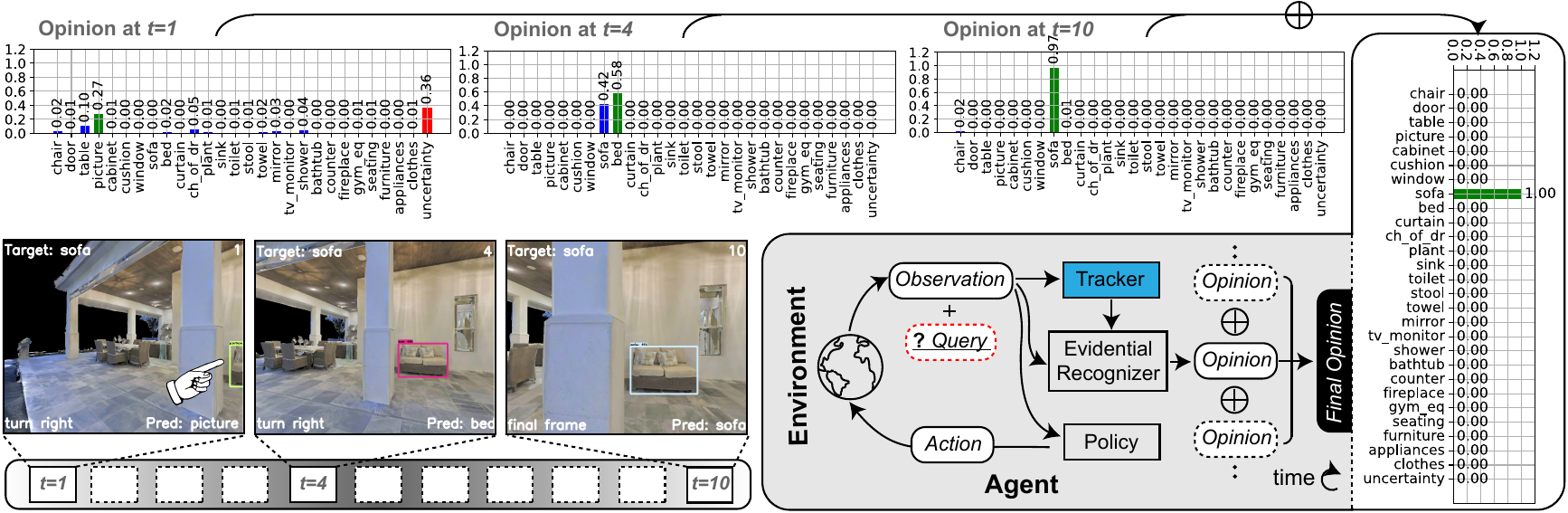}
    \caption{An illustration of a recognition episode and the proposed agent's architecture in the grey box. We select three frames ($t=1,4,10$) along with their estimated opinions. The bars for top prediction and uncertainty are colored green and red, respectively. Note that uncertainty arises when the target is partially out of view at the first step. Despite this, the final result is accurate due to the fusion of evidence.}
    \vspace{-4pt}
    \label{fig:method}
\end{figure*}

The architecture of proposed agent is demonstrated in Figure.~\ref{fig:method}.
Besides the recognition model to predict the category of the target, the policy module $\pi_{\phi}(a^t|s^{t-1})$ with parameters $\phi$ is supposed to control the robot to maximize the accumulated reward $R=\sum_{t=2}^T r^t$.
The state $s^{t-1}$ describes the aggregated information from observations till the timestep $t-1$.
To prevent interference with the recognition, we use a separate recurrent unit $g(\cdot)$ for fusing temporal knowledge.
The state is then expressed as $s^{t}=g(v^1,\dots,v^{t},q_x^t)$, where $q_x^t$ denotes the binary mask of the target $x$ at timestep $t$.
To ensure clarity in our study and to eliminate uncertainties introduced by other modules, the mask $q_x^t$ is derived from the ground-truth bounding box of the target, consistent with previous embodied perception works~\cite{zhao2023zero,wortsman2019learning,pal2021learning}. 
In practical applications, the agent can utilize off-the-shelf class-agnostic visual trackers~\cite{wang2019fast} to obtain $q_x^t$.

We design a novel reward to stabilize policy learning for ambiguous visual observations.
According to our prediction result as in Equation~\ref{eq:belief_unc}, the reward $r^t$ should reflect the evidence of the correct class.
To this end, the proposed reward is straightforwardly defined as $r^t=b^t_y$.
Note that we use $b^t_y$ to represent the belief for the correct class, which can also be interpreted as the normalized evidence.
The reward $r^t$ varies between 0 to 1 depending on how much the recognition model collects the evidence for the target class.
Since our output includes an uncertainty term, the belief for all known classes may be extremely low when encountering an unexpected input, suggesting that the action leading to the observation should not be rewarded.

As discussed in~\cite{fan2023avoiding,yang2019embodied}, joint training of the recognition module and the policy can lead to a collapsed outcome. 
This is because an insufficiently-trained recognition model may not provide accurate rewards during the early training stages. 
Therefore, we perform a staged training: first, we train the recognition model using a heuristic policy, and then we train only the policy part using PPO~\cite{schulman2017proximal}, while keeping the recognition model fixed.


\section{Testing Dataset for Active Recognition}
\label{sec:dataset}

\begin{figure}[t]
    \centering
        \includegraphics[width=1\linewidth]{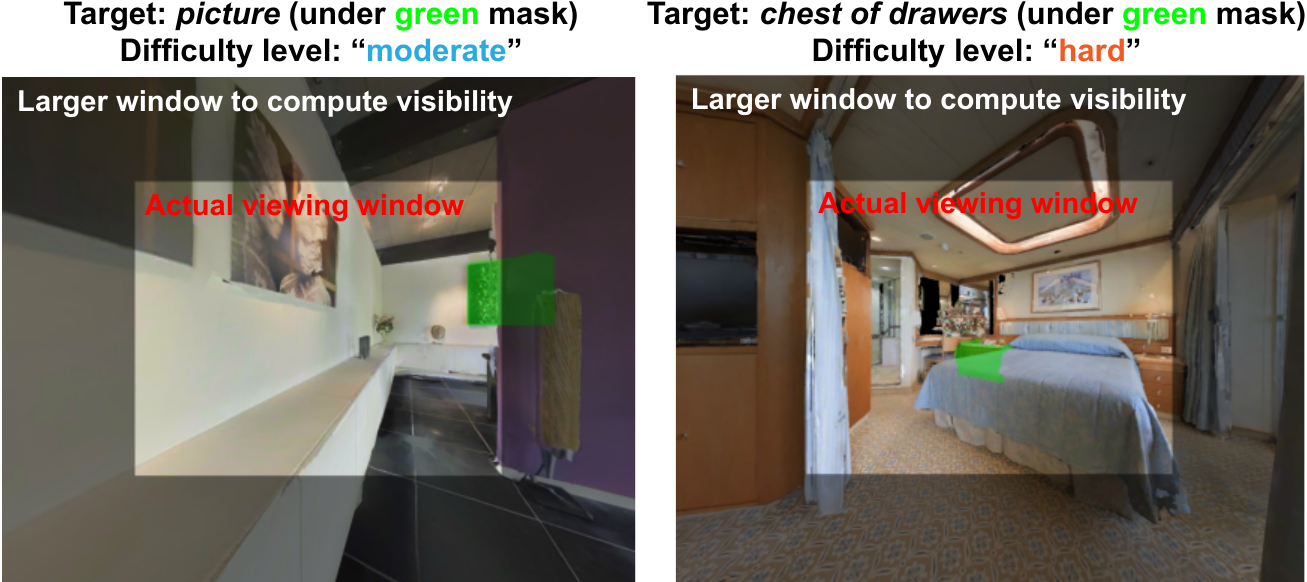}
        \includegraphics[width=0.9\linewidth]{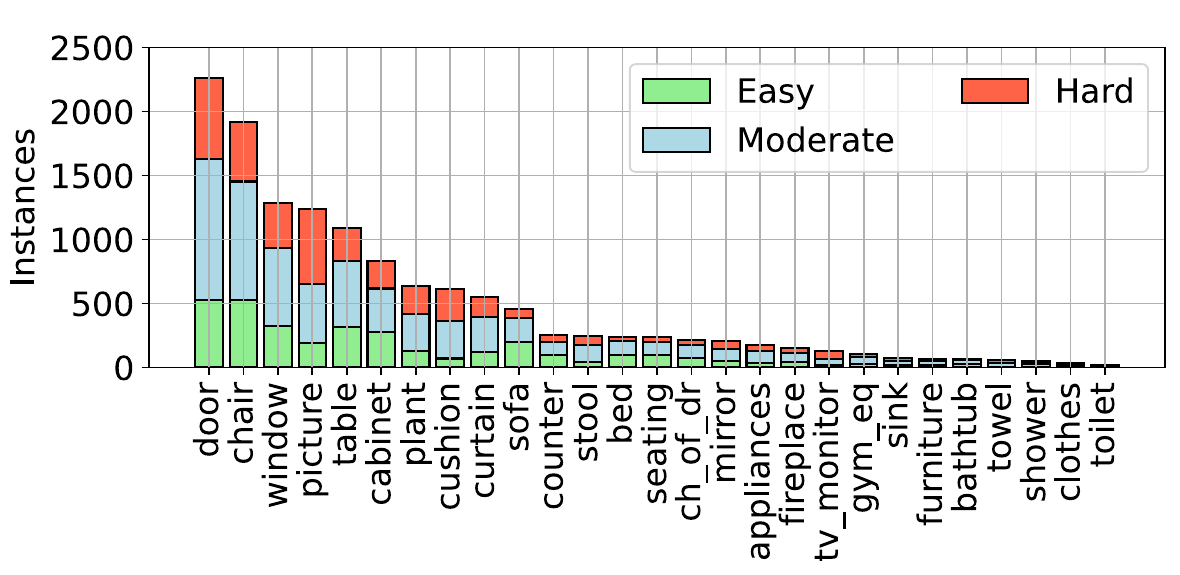}
    \caption{We present two examples of test data from the proposed dataset in the upper sub-figure. The target object is covered by a green mask, allowing for box, point or mask queries during testing. The targets are respectively occluded by the wall and the bed, thus visibility is calculated as the ratio of observed pixels to total pixels belonging to the target. In the lower sub-figure, the distribution of difficulty levels across all categories is depicted.}
    \vspace{-8pt}
    \label{fig:dataset}
\end{figure}
Compared to passive recognition, active recognition is motivated to avoid undesired viewing conditions, including but not limited to heavy occlusions and the out-of-view, where adjusting viewpoints is inevitable.
We build a testing dataset designated for assessing the performance of active recognition under different recognition challenges, which is collected from the current simulator MP3D~\cite{chang2017matterport3d,savva2019habitat,szot2021habitat}. 

The testing dataset from MP3D~\cite{chang2017matterport3d} contains 90 building-scale indoor spaces covering diverse scene categories. 
For each test instance, the agent's starting position and the query are randomly specified, ensuring that the relative distance between them falls within the range of 3 to 6 meters.
Since the semantic annotation in MP3D~\cite{chang2017matterport3d} could be noisy, we manually remove erroneous or low-quality queries, resulting in a total of $13200$ testing instances.
The training of the agent is conducted using the non-overlapping HM3D dataset~\cite{ramakrishnan2021hm3d}, which contains 145 semantically-annotated indoor scenes, along with randomly generated initial positions and queries.


\noindent{\bf{Object categories.}} We select $27$ object categories from over $40$ labeled classes that exists in both datasets~\cite{chang2017matterport3d,ramakrishnan2021hm3d}. A category is selected only if it does not contain high ambiguities (objects, misc, {\it{etc}}.) or belongs to building components (wall, ceiling, stairs, {\it{etc}}.).

\noindent{\bf{Recognition difficulty.}} We introduce an estimate of recognition difficulty for each testing instance. Unlike~\cite{yang2019embodied}, we evaluate the difficulty level from three perspectives: {\it{visibility}}, {\it{relative distance}}, and {\it{observed pixels}}. {\it{Visibility}} is determined by the ratio computed by dividing the unoccluded segmentation mask of the target by its observed mask. 
To consider partially out-of-view conditions, the unoccluded mask is captured within an enlarged viewing window.
Two examples of test instances are depicted at the top of Figure~\ref{fig:dataset}, where the left target object is not entirely within the agent's viewing window, and the right target is additionally occluded by a bed.
The second aspect, {\it{relative distance}}, measures the distance between the agent and the target object.
This is determined by subtracting the normalized relative distance (originally ranging from 3 to 6 meters) from 1.
{\it{Observed pixels}} account for the visible pixels of the target object and are normalized with a cap at 102400 pixels.
The final difficulty score combines these three factors using weights of 0.2, 0.2, and 0.6, respectively. We emphasize {\it{observed pixels}} because the other two aspects might not sufficiently represent the recognition challenges posed by tiny objects.

The difficulty level is assigned as follows: "hard" if the score is less than 0.33, "moderate" if it falls between 0.33 and 0.66, and "easy" for all other cases.
The difficulty level for each category is depicted in the lower part of Figure.~\ref{fig:dataset}.
Since we do not re-balance or re-sample rare classes, the category distribution follows a long-tail pattern, reflecting their true occurrences in testing scenes.
The dataset will be made publicly available to facilitate reproducible research comparisons.
More detailed dataset generation and statistics can be found in the supplementary.


\section{Experiments}
\label{sec:experiment}
To validate the effectiveness of our method, we first compare the proposed method with other approaches using our new dataset, demonstrating the benefits of uncertainty-aware policy learning and evidence fusion.
Next, we examine the behavior of provided uncertainties across various dimensions, such as categories, steps, and recognition difficulties.
More experiments, including ablation studies, are conducted in the final part to further evaluate our method.

\subsection{Implementations}
The recognition model, based on Faster R-CNN~\cite{ren2015faster}, is modified by replacing the region proposal network with the query box at the current timestep.
The backbone is ResNet-50~\cite{he2016deep} pretrained on the ImageNet~\cite{russakovsky2015imagenet}.
We use the ROI features with a C4 head~\cite{ren2015faster} and also fix the first three residual blocks during training as in~\cite{yang2019embodied}.
The training of the recognition model incorporates a heuristic fixation coupled with a shortest-path policy, which attempts to approach the target while positioning it at the center of the view.
Furthermore, the target is directly provided by the ground-truth bounding box at all timesteps.

The policy part is trained with the recognition model held fixed.
The policy network comprises an independent visual encoder, a single-layer Gated
Recurrent Unit (GRU) to integrate temporal knowledge, and two single linear-layer to perform as actor and critic on the GRU's output. 
For detailed architecture and other training hyper-parameters, please refer to our supplementary.

\subsection{Baselines}
The compared baselines serve two purposes, \ie, to demonstrate the effectiveness of incorporating intelligent strategies into the recognition process, and to highlight the advantages of modeling uncertainties in active recognition.
Heuristic recognition policy baselines, including {\texttt{Single-View}}, {\texttt{Random}}, and {\texttt{Fixation}}, employ a similar network architecture, but with alterations to the policy part.

\noindent{\texttt{Single-View}}: It imitates conventional passive recognition models that could not intelligently perceiving the target. It uses the initial observation for prediction.

\noindent{\texttt{Random}}: The policy randomly selects an action at each step. The number of
movements remains the same as ours.

\noindent{\texttt{Fixation}}: This policy aims to center the target within the view, potentially reducing undesired out-of-view conditions.

\noindent{\texttt{Amodal-Rec}}: Furthermore, we re-implement the embodied amodel recognition agent proposed in~\cite{yang2019embodied} which uses an additional convolutional GRU to recurrently aggregate visual features and generate predictions using softmax probabilities.
To ensure a fair comparison, we replace the training method in {\texttt{Amodal-Rec}}~\cite{yang2019embodied} from REINFORCE~\cite{sutton1998introduction} to PPO~\cite{schulman2017proximal} and also the perception part from Mask R-CNN~\cite{he2017mask} to Faster R-CNN~\cite{ren2015faster}, aligning it with our approach.


\subsection{Performance Evaluation}

\begin{table*}[t]
\centering
\footnotesize
\caption{
Recognition success rates and improvements on the proposed dataset.
Success rates are measured according to the final predictions, while the changes in success rates  highlight the improvements achieved through movements, thus demonstrating the effectiveness of various policies.
{\texttt{Random}} and {\texttt{Fixation}} are two heuristic policies that can be integrated with different recognition agents.
}
\label{tab:main}
\begin{tabular}{c|cl|cl|cl|cc|cccc}
\hline
\multirow{2}{*}{\textbf{Method}} & \multicolumn{2}{c|}{\textbf{Easy}} & \multicolumn{2}{c|}{\textbf{Moderate}} & \multicolumn{2}{c|}{\textbf{Hard}} & \multicolumn{2}{c|}{\textbf{All}} & \multicolumn{4}{||c}{\textbf{Change in success rate ($t=1$ to $10$)}} \\ \cline{2-13} 
 & \multicolumn{1}{c|}{\textit{top-1}} & {\textit{top-3}} & \multicolumn{1}{c|}{\textit{top-1}} & {\textit{top-3}} & \multicolumn{1}{c|}{\textit{top-1}} & {\textit{top-3}} & \multicolumn{1}{c|}{\textit{top-1}} & \textit{top-3} & \multicolumn{1}{||c|}{{\textbf{Easy}}} & \multicolumn{1}{c|}{{\textbf{Moderate}}} & \multicolumn{1}{c|}{{\textbf{Hard}}} & {\textbf{All}} \\ \hline\hline
{\texttt{Amodal-Rec}} + {\texttt{Single-View}} & \multicolumn{1}{c|}{65.1} & 84.3 & \multicolumn{1}{c|}{55.3} & 77.1 & \multicolumn{1}{c|}{43.3} & 66.5 & \multicolumn{1}{c|}{57.4} & 78.0 & \multicolumn{1}{||c|}{-} & \multicolumn{1}{c|}{-} & \multicolumn{1}{c|}{-} & - \\ \hline
{\texttt{Amodal-Rec}} + {\texttt{Random}}  & \multicolumn{1}{c|}{64.7} & 83.2 & \multicolumn{1}{c|}{55.2} & 76.8 & \multicolumn{1}{c|}{42.4} & 66.3 & \multicolumn{1}{c|}{57.0} & 77.6 & \multicolumn{1}{||c|}{-0.4} & \multicolumn{1}{c|}{-0.1} & \multicolumn{1}{c|}{-0.9} & -0.4 \\ \hline
{\texttt{Amodal-Rec}} + {\texttt{Fixation}} & \multicolumn{1}{c|}{64.8} & 83.3 & \multicolumn{1}{c|}{55.1} & 76.7 & \multicolumn{1}{c|}{44.4} & 67.4 & \multicolumn{1}{c|}{57.1} & 77.6 & \multicolumn{1}{||c|}{-0.3} & \multicolumn{1}{c|}{-0.2} & \multicolumn{1}{c|}{+1.1} & -0.3 \\ \hline
{\texttt{Amodal-Rec}} & \multicolumn{1}{c|}{65.0} & 83.5 & \multicolumn{1}{c|}{55.1} & 76.6 & \multicolumn{1}{c|}{42.9} & 66.5 & \multicolumn{1}{c|}{57.2} & 77.7 & \multicolumn{1}{||c|}{-0.1} & \multicolumn{1}{c|}{-0.2} & \multicolumn{1}{c|}{-0.4} & -0.2 \\ \hline\hline
{Ours + \texttt{Single-View}} & \multicolumn{1}{c|}{67.9} & 86.2 & \multicolumn{1}{c|}{57.1} & 77.8 & \multicolumn{1}{c|}{49.7} & 70.8 & \multicolumn{1}{c|}{60.9} & 80.7 & \multicolumn{1}{||c|}{-} & \multicolumn{1}{c|}{-} & \multicolumn{1}{c|}{-} & - \\ \hline
Ours + {\texttt{Random}} & \multicolumn{1}{c|}{62.0} & 86.0 & \multicolumn{1}{c|}{49.0} & 74.9 & \multicolumn{1}{c|}{39.2} & 67.6 & \multicolumn{1}{c|}{53.4} & 78.8 & \multicolumn{1}{||c|}{-5.9} & \multicolumn{1}{c|}{-8.1} & \multicolumn{1}{c|}{-10.5} & -7.5 \\ \hline
Ours + {\texttt{Fixation}} & \multicolumn{1}{c|}{66.2} & 88.0 & \multicolumn{1}{c|}{58.0} & 79.0 & \multicolumn{1}{c|}{56.1} & 78.3 & \multicolumn{1}{c|}{61.8} & 83.7 & \multicolumn{1}{||c|}{-1.7} & \multicolumn{1}{c|}{+0.9} & \multicolumn{1}{c|}{+6.4} & +0.9 \\ \hline
Ours & \multicolumn{1}{c|}{{\textbf{69.9}}} & {\textbf{88.3}} & \multicolumn{1}{c|}{{\textbf{59.7}}} & {\textbf{80.4}} & \multicolumn{1}{c|}{{\textbf{58.0}}} & {\textbf{80.2}} & \multicolumn{1}{c|}{{\textbf{64.4}}} & {\textbf{84.3}} & \multicolumn{1}{||c|}{{\textbf{+2.0}}} & \multicolumn{1}{c|}{{\textbf{+2.6}}} & \multicolumn{1}{c|}{{\textbf{+8.3}}} & {\textbf{+3.5}} \\ \hline
\end{tabular}
\vspace{-4pt}
\end{table*}


In this section, we present a quantitative comparison across active recognition models under various evaluation metrics.
In line with established protocols~\cite{jayaraman2016look,yang2019embodied}, we set a limit of $T=10$ total allowed steps and subsequently report the performance at the final step.
We provide the success rate across varying levels of recognition difficulty.
Furthermore, we present a comparative analysis of the success rate from the initial to the final step to illustrate the advantages gained by integrating various movement policies.

In Table.~\ref{tab:main}, we observe that the proposed method achieves improvement over heuristic polices on testing instances with different difficulty levels.
For example, the success rate of proposed method is $2.6\%$ higher than with {\texttt{Fixation}} policy, while $9.0\%$ higher for with {\texttt{Random}} policy.
This is primarily because the {\texttt{Random}} policy is prone to losing sight of the target easily, while the {\texttt{Fixation}} policy does not contribute significant information once the target appears at the center of the view.

Another key observation relates to the results associated with "hard" recognition instances. 
The increase in success rate for the proposed method on "hard" testing instances ($+8.3\%$) is significantly greater than that on "easy" ($+2.0\%$) and "moderate" ($+2.6\%$) instances. 
This improvement likely stems from the fact that "hard" instances often involve substantial occlusions and greater distances, conditions that require the agent to adopt more sophisticated viewing strategies in order to acquire clear and distinct observations. 
We provide a detailed analysis of visibility and distance variations at each step in the supplementary materials.

\begin{figure}[t]
    \centering
    \includegraphics[width=0.8\linewidth]{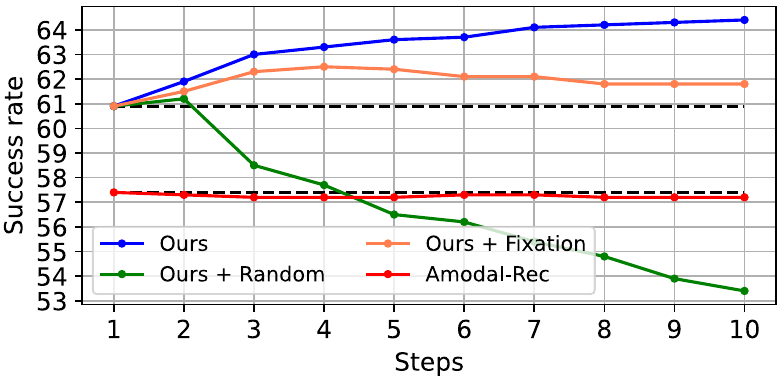}
    \vspace{-6pt}
    \caption{Performance between different agents over steps.}
    \vspace{-2pt}
    \label{fig:succ_by_step}
\end{figure}

Furthermore, the aggregation of temporal information at the feature level encounters challenges when deployed in open-world environments. This finding stems from a comparison of the results derived from {\texttt{Amodal-Rec}}~\cite{yang2019embodied} with a trained policy versus the corresponding {\texttt{Single-View}} approach.
Step-by-step results for {\texttt{Amodal-Rec}}~\cite{yang2019embodied} are depicted in Figure~\ref{fig:succ_by_step}, revealing a steady trend in recognition as more steps are taken. 
We postulate two potential reasons for this phenomenon.
Firstly, {\texttt{Amodal-Rec}}~\cite{yang2019embodied} collates whole-image features across varying frames and subsequently leverages the initial query location to pool regional information from the aggregated feature for classification. 
As the agent alters its viewpoints, the regional feature may lose its discriminative power as the same queried region incorporates irrelevant information in subsequent frames.
Secondly, our dataset presents greater challenges compared to the one used in {\texttt{Amodal-Rec}}~\cite{yang2019embodied}, primarily due to increased freedom of movement, \ie, the ability to look up and down. 
Consequently, the likelihood of the same queried region pertaining to the same object between consecutive frames diminishes.
These observations underscore the jeopardy of overlooking the inherent challenges posed by active recognition in open-world environments.


The success rate over steps is depicted in Figure~\ref{fig:succ_by_step}. 
Overall, the proposed method shows an improvement in performance as more observations are taken, eventually reaching a saturation point.
Interestingly, there is a noticeable performance surge at $t=2$ for our method with {\texttt{Random}}. 
This can be attributed to the high probability of the target existing within the observation after a single movement. 
In essence, the proposed evidence fusion strategy can enhance recognition if the target is observable. Moreover, significant improvements can be achieved through the implementation of our uncertainty-aware policy learning approach.

\subsection{Uncertainty in Active Recognition}
\begin{figure}[t]
    \centering
    \includegraphics[width=1\linewidth]{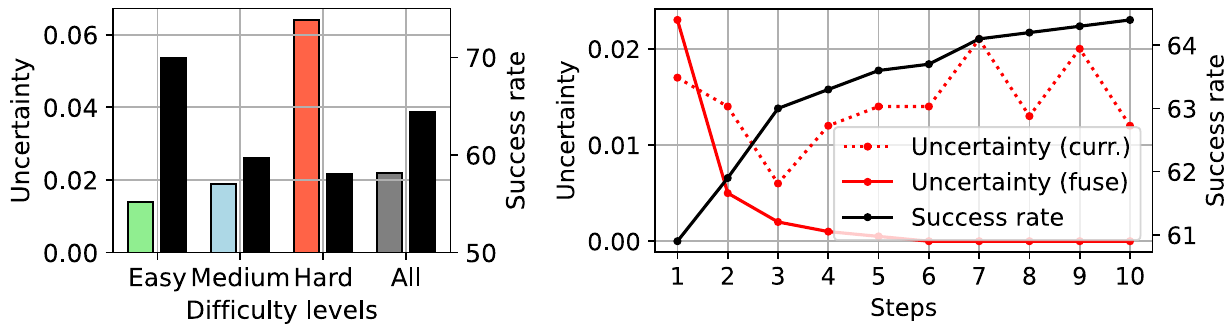}
    \caption{The left illustrates the variation in average uncertainties across different levels of difficulty, with each pair of bars representing the uncertainty and the corresponding success rate. The right section examines the changes in uncertainty.}
    \vspace{-4pt}
    \label{fig:uncertainty}
\end{figure}

\begin{figure}[t]
    \centering
    \includegraphics[width=1\linewidth]{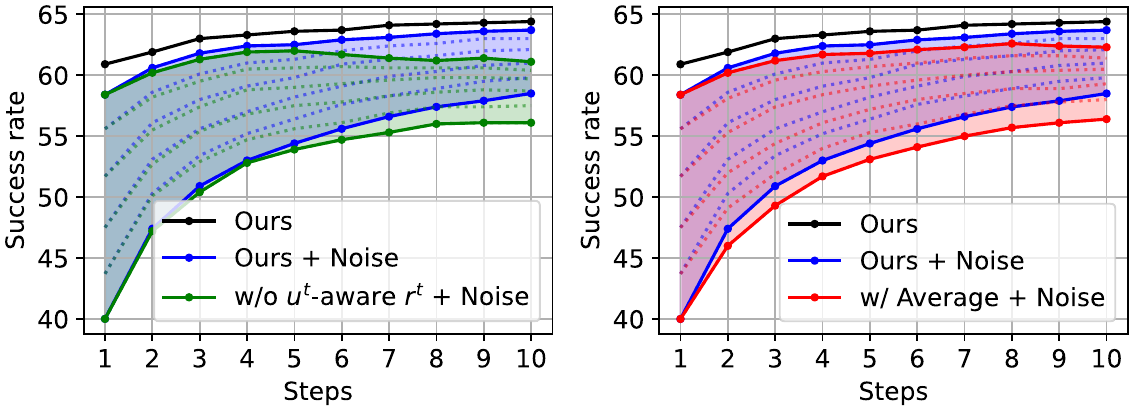}
    \caption{Step-by-step success rates of different variants of the proposed method under varying levels of feature noise. Six noise levels, modeled as Gaussian with $\mu=0$ and $\sigma=\{2,3,4,5,6,7\}$, are added to the features produced by our fixed backbone.}
    \vspace{-8pt}
    \label{fig:perturbance}
\end{figure}
In this section, we delve further into the behavior of uncertainties in our approach. The uncertainty and corresponding final success rate across different difficulty levels are jointly depicted on the left of Figure.~\ref{fig:uncertainty}.
For clarity, the uncertainty depicted in the left sub-figure represents the average of the uncertainties at each step, aggregated across all test instances.
During testing, as the agent actively acquiring discriminative views for recognition, the uncertainty generally remains at a low level.
Furthermore, it can be observed that the uncertainty increases with higher recognition challenges, indicating its potential ability to discern challenging recognition queries.

The trends of step-by-step fused uncertainties and success rates are reported on the right side of Figure.~\ref{fig:uncertainty}. The fused uncertainty decreases and recognition improves as more observations are taken, indicating that the proposed agent effectively gathers evidence to support prediction-making. 
Additionally, we include the uncertainty before fusion for reference (represented by the dotted red curve), which is independently predicted based on the current view.

\subsection{Performance under Feature Disturbance}
In this section, we examine the impact of our uncertainty-aware reward strategy during training, and our evidence fusion method, on performance in varied environments. 
Due to it is impractical and unpredictable to manipulate simulation environments to control recognition challenges, we opted to add Gaussian noise into the visual features obtained by the pre-trained ResNet-50 backbone~\cite{he2016deep} to simulate unexpected recognition scenarios. 
These perturbed features are subsequently input into the classification component for per-step prediction, culminating in a fused final output.

We specifically implemented six levels of Gaussian noise, characterized by a mean ($\mu=0$) and standard deviations ($\sigma=\{2,3,4,5,6,7\}$).
A higher $\sigma$ value signifies more intense feature perturbations, presenting greater recognition challenges.
Our comparative analysis included two variants of our method, each utilizing the same model architecture.
The step-by-step success rates are depicted in Figure~\ref{fig:perturbance}.

Initially, we evaluated the performance of our method in the absence of uncertainty-aware rewards $r^t$ during policy learning.
In this configuration, a binary reward was assigned to the policy based on its current prediction.
As illustrated in the left of Figure~\ref{fig:perturbance}, our approach demonstrated enhanced robustness against feature noise compared to the model trained without uncertainty-aware rewards. 
We argue that a binary reward system fails to adequately represent the value of actions during training, particularly under conditions of high visual uncertainty, leading to a less effective policy.

Additionally, we substituted our proposed evidence fusion method with an {\it{Average}} method, which calculates the mean of estimated beliefs across all frames before making a prediction.
This result is showcased in the right of Figure~\ref{fig:perturbance}.
The {\it{Average}}, not accounting for uncertainties in each frame, results in an integrated feature that lacks discriminative power, especially in scenarios with high feature noise.



\subsection{Ablation Studies}

\begin{table}[t]
\centering
\scriptsize
\caption{Ablation studies about different evidence fusion strategies and the integration of uncertainty-aware reward.}
\label{tab:ablation}
\begin{tabular}{c|c|c|c|cc}
\hline
\multirow{2}{*}{\textbf{Method}} & \textbf{Easy} & \textbf{Moderate} & \textbf{Hard} & \multicolumn{2}{c}{\textbf{All}} \\ \cline{2-6} 
 & {\textit{top-1}} & {\textit{top-1}} & {\textit{top-1}} & \multicolumn{1}{c|}{{\textit{top-1}}} & {\textit{top-3}} \\ \hline\hline
Ours & {\textbf{69.9}} & {\textbf{59.7}} & {\textbf{58.0}} & \multicolumn{1}{c|}{{\textbf{64.4}}} & {\textbf{84.3}} \\ \hline
w/ {\it{Max-prediction}} & 69.3 & 56.8 & 56.4 & \multicolumn{1}{c|}{62.9} & 80.3 \\ \hline
w/ {\it{Last-step}} & 66.2 & 55.2 & 55.7 & \multicolumn{1}{c|}{60.7} & 79.5 \\ \hline
w/ {\it{Average}} & 68.5 & 57.8 & 57.5 & \multicolumn{1}{c|}{63.0} & 82.6 \\ \hline
w/ {\it{Vote}} & 68.6 & 57.8 & 56.3 & \multicolumn{1}{c|}{62.8} & 83.1 \\ \hline\hline
w/o $u^t$-aware $r^t$ & 67.5 & 57.8 & 53.4 & \multicolumn{1}{c|}{61.6} & 81.3 \\ \hline
\end{tabular}
\vspace{-5pt}
\end{table}

In this study, we explore the impact of different factors such as evidence fusion methods and training rewards on our performance. 
To eliminate any unrelated interference, we employ the same visual recognition and policy model for all compared methods discussed in this section.

We first evaluate the proposed evidence fusion method against four alternative strategies: {\it{Max-prediction}}, {\it{Last-step}}, {\it{Average}}, and {\it{Vote}}. 
The {\it{Max-prediction}} strategy selects the prediction with the highest estimated belief from all steps as the final output, indicating the highest model confidence at that particular step. 
The {\it{Last-step}} strategy solely considers the final single-frame estimation as the outcome. 
Lastly, the {\it{Vote}} strategy implements a voting mechanism among all frame predictions to determine the outcome. 
Table~\ref{tab:ablation} presents the success rates for various fusion strategies. 
Our method surpasses these strategies across all levels of recognition difficulty. 
This superiority primarily stems from the fact that the four alternative strategies fail to account for potential uncertainties arising at each step of embodied recognition, such as occlusions. 
As detailed in Equation~\ref{eq:fusion}, our approach accumulates frame-wise evidence while factoring in estimated uncertainties; thereby, estimates with higher uncertainty exert less influence on the final prediction.

Furthermore, the comparison with agent trained without uncertainty-aware reward emphasizes the efficacy of the proposed reward during training, especially for "hard" testing instances that contain high ambiguities. 
Essentially, this also validates the importance of managing challenges inherent in open environments.

\subsection{Limitations and Future Works}
In our experiments, recognition performance is evaluated with a predetermined number of total steps, yet an ideal agent would be able to determine when to cease taking more movements. 
Accordingly, as both recognition accuracy and the number of visual observations vary, an effective evaluation metric is further required.

\section{Conclusions}
\label{sec:conclusion}
In this paper, we examine the challenges faced when deploying active recognition agents in open-world environments, specifically, how to avoid negative impacts from unexpected inputs in class prediction and policy learning.
These challenges are inherent in active recognition due to the unpredictable and open nature of the environments being explored. 
To address this, we propose to model the recognition as a sequential evidence-collecting process, leading to an uncertainty-aware agent.
Observations from unknown classes or highly ambiguous views can be rejected, fostering more stable and effective policy learning.
A reduced hypothesis space is introduced for evidence fusion, generating the final opinion in accordance with evidence combination theory. 
A new dataset annotated with recognition difficulties is introduced to evaluate different agents.
Experiments on the dataset, along with the uncertainty analyses and ablation studies, confirm the effectiveness of our proposed method.

{
    \small
    \bibliographystyle{ieeenat_fullname}
    \bibliography{main}
}


\end{document}